\documentclass[10pt,twocolumn,letterpaper]{article}

\usepackage{cvpr}              

\definecolor{cvprblue}{rgb}{0.21,0.49,0.74}
\usepackage[pagebackref,breaklinks,colorlinks,allcolors=cvprblue]{hyperref}
\usepackage[dvipsnames]{xcolor}
\usepackage{multirow}
\usepackage{graphicx}
\usepackage{subcaption}
\usepackage{rotating}
\usepackage[accsupp]{axessibility}  
\def\paperID{7895} 
\def\confName{CVPR}
\def\confYear{2025}

\title{Link to the Past:  \textcolor{CornflowerBlue}{Tem}poral \textcolor{CornflowerBlue}{P}r\textcolor{CornflowerBlue}{o}pagation for \textcolor{CornflowerBlue}{Fast 3D} Human Reconstruction from Monocular Video}

\author{First Author\\
Institution1\\
Institution1 address\\
{\tt\small firstauthor@i1.org}
\and
Second Author\\
Institution2\\
First line of institution2 address\\
{\tt\small secondauthor@i2.org}
}

\author{Matthew Marchellus, Nadhira Noor, and In Kyu Park\\
Department of Electrical and Computer Engineering, Inha University\\
Incheon 22212, Korea\\
{\tt\small \{marchellusmatthew@gmail.com, nadhirannoor@gmail.com, pik@inha.ac.kr\}}
}

\begin{document}

\twocolumn[{
\maketitle

\begin{center}
    \includegraphics[width=\textwidth]{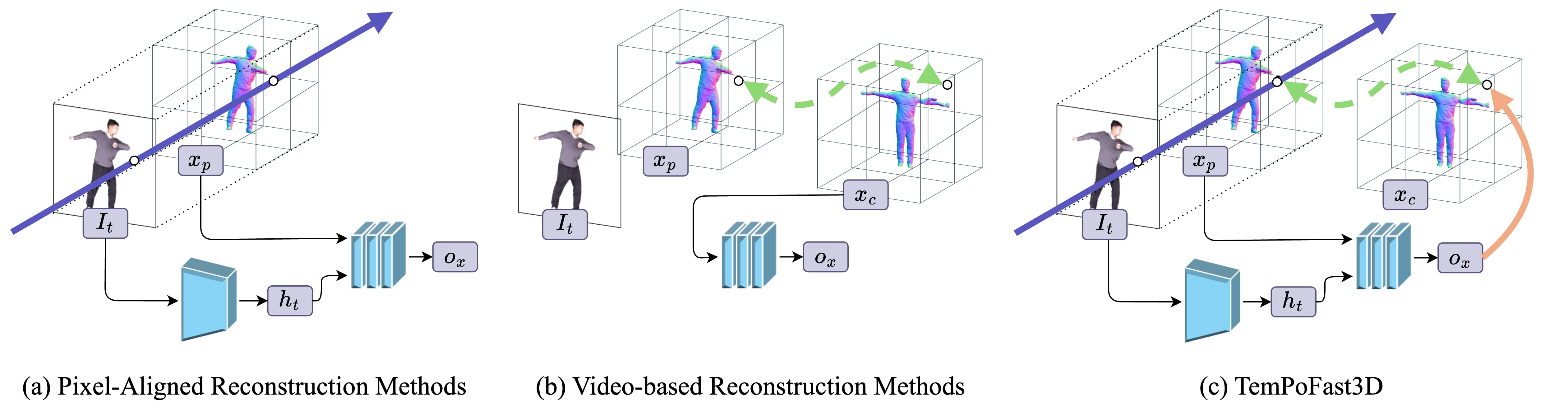} 


    
    \captionof{figure}{We propose \textbf{TemPoFast3D}, a novel pipeline to leverage the temporal coherency of human appearance for efficient and accurate 3D human reconstruction from monocular videos. We temporally propagate information from the past frames result by blending the pixel-aligned implicit function and avatar reconstruction method.}
    \vspace*{2mm}
    \label{fig:banner_figure}
\end{center}
}]


\begin{abstract}
Fast 3D clothed human reconstruction from monocular video remains a significant challenge in computer vision, particularly in balancing computational efficiency with reconstruction quality.
Current approaches are either focused on static image reconstruction but too computationally intensive, or achieve high quality through per-video optimization that requires minutes to hours of processing, making them unsuitable for real-time applications.
To this end, we present TemPoFast3D, a novel method that leverages temporal coherency of human appearance to reduce redundant computation while maintaining reconstruction quality.
Our approach is a ``plug-and play" solution that uniquely transforms pixel-aligned reconstruction networks to handle continuous video streams by maintaining and refining a canonical appearance representation through efficient coordinate mapping.
Extensive experiments demonstrate that TemPoFast3D matches or exceeds state-of-the-art methods across standard metrics while providing high-quality textured reconstruction across diverse pose and appearance, with a maximum speed of 12~FPS.

\end{abstract}


\section{Introduction}
\label{sec:introduction}

Real-time 3D human reconstruction from monocular video streams is a fundamental challenge that could revolutionize virtual reality, telepresence, and human-computer interaction.
These applications demand methods that can accurately reconstruct 3D clothed humans from single images or video streams, capturing both detailed geometry and realistic appearance. 
However, existing approaches often struggle to balance the computational efficiency required for video with the high fidelity demanded by real-world applications.

Recent approaches to 3D clothed human reconstruction primarily follow two distinct paradigms. 
(i) Single-image reconstruction methods rely on pixel-aligned features~\cite{pifu2019Saito, pifu-hd2020Saito, pamir2022Zheng, phorhum2022Alldieck, geo-pifu2020He, icon2022Xiu, econ2023Xiu, gta2023Zhang} to capture detailed geometry and textures through implicit functions.
(ii) Video-based reconstruction methods leverage pose deformation with either implicit neural fields~\cite{HumanNerf2022Weng, Vid2Avatar2023Guo, InstantAvatar2023Jiang, NeuMan2022Jiang, SelfRecon2022Jiang, NeuralBody2021Peng, Anerf2021Su, Danbo2022Su} or Gaussian splatting~\cite{GaussianAvatar2024Hu,3dgsAvatar2024qian,X-Avatar2023shen,ExAvatar2024moon} for articulated 3D modeling.
While producing high-quality results, these methods require extensive optimization ({\it i.e.} minutes to hours) and multiple passes over the video, making them unsuitable for real-time applications.
Recent attempts at real-time reconstruction~\cite{fof2022Feng} either sacrifice quality, require additional inputs (templates~\cite{LiveCap2019Habermann, MonoPerfCap2018Xu}, multi-view~\cite{multiview12015collet, multiview22021Zhang}, depth~\cite{depth42011newcombe, depth22015Newcombe, depth32022Dong}), or lack true 3D capability~\cite{Monoport2020Li, fof2022Feng, rtvrendering2023rocco}, leaving fast, high-quality reconstruction from video an unsolved challenge.

Our key insight is that while human poses change rapidly across video streams, the underlying body shape and clothing geometry remain largely consistent over short time periods.
Video-based methods exploit this temporal coherence, but require multiple passes over the entire sequence for global optimization.
We observe that this temporal consistency can be leveraged even further for faster reconstruction through progressive canonical shape learning for sequential frame-by-frame processing.
However, the challenge of learning a canonical shape through sequential frame-by-frame processing remains largely unexplored in existing literature.
We address this limitation by combining the reconstruction accuracy of pixel-aligned methods~\cite{pifu2019Saito, pifu-hd2020Saito, pamir2022Zheng, phorhum2022Alldieck, geo-pifu2020He, icon2022Xiu, econ2023Xiu, gta2023Zhang} with bidirectional canonical-posed space mapping~\cite{HumanNerf2022Weng, Vid2Avatar2023Guo, InstantAvatar2023Jiang, NeuMan2022Jiang, SelfRecon2022Jiang, NeuralBody2021Peng, Anerf2021Su, Danbo2022Su}, enabling efficient shape learning while maintaining high reconstruction quality from video streams.

To that end, we propose \textbf{TemPoFast3D}, a novel fast frame-by-frame 3D reconstruction approach that temporally propagates canonical shape information across video frames.
Our method combines pixel-aligned reconstruction with SMPL-based coordinate mapping to maintain a consistent canonical representation while accurately capturing pose variations.
This combination naturally extends to multi-view settings when additional views are available.
Furthermore, we develop optimization strategies including adaptive coordinate sampling and visibility-guided filtering that significantly reduce per-frame computation.
Our framework is designed as a ``plug-and-play" solution that accelerates existing SMPL-guided pixel-aligned reconstruction methods, reaching maximum speed of 12 frames per second while maintaining reconstruction quality.
In summary, our contributions are:
\begin{itemize}
\setlength\itemsep{0.2em}
\item \begin{minipage}[t]{\linewidth} We introduce TemPoFast3D, a novel framework combining canonical space inference with efficient pose deformation, enabling faster 3D clothed human reconstruction from monocular video without additional inputs or templates\end{minipage}
\item \begin{minipage}[t]{\linewidth} A suite of optimization strategies including adaptive coordinate sampling and visibility-guided filtering mechanisms that significantly reduce per-frame computation while preserving reconstruction quality\end{minipage}
\item \begin{minipage}[t]{\linewidth}A plug-and-play design that accelerates existing SMPL-aligned reconstruction methods while enabling additional capabilities such as multi-view reconstruction for enhanced quality\end{minipage}
\end{itemize}


\begin{figure*}[t]
  \centering
   \includegraphics[width=1\linewidth]{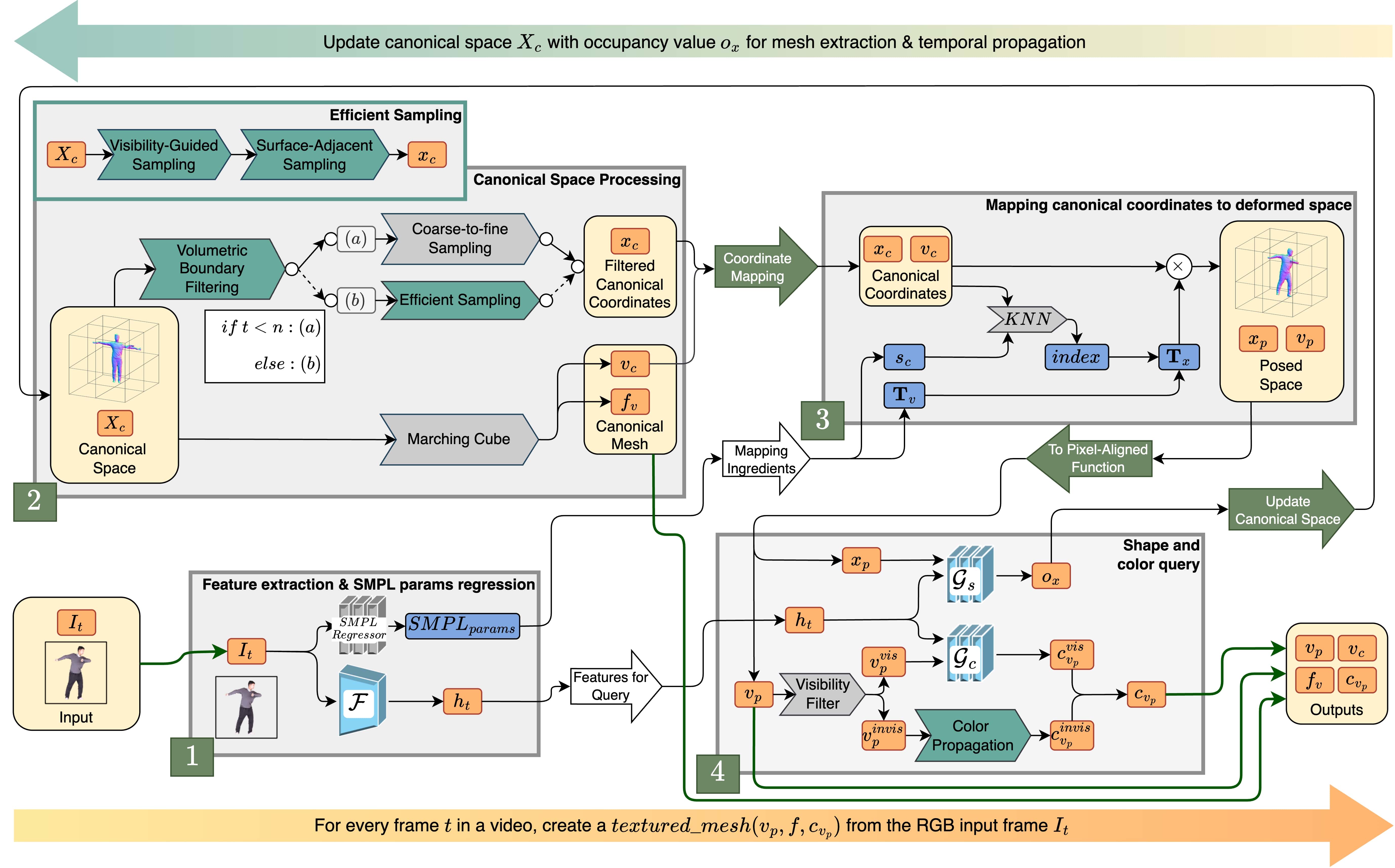}
   \caption{\textbf{Overview of our TemPoFast3D pipeline.} Given an input RGB frame $I_t$, our method combines efficient canonical space processing with coordinate mapping for fast 3D human reconstruction. The pipeline consists of: (Section~\ref{ssec:preliminary}) Feature extraction and SMPL params regression, (Section~\ref{ssec:canspace_inference}) Mapping canonical coordinates to posed space, (Section~\ref{pg:posed}) Shape and color query, and (Section~\ref{ssec:temp_pro}) Canonical space processing. The canonical space representation $X_c$ is continuously updated across frames.}
   \label{fig:main_pipeline}
\end{figure*}

\section{Related Works}
\label{sec:related_works}

\noindent\textbf{Pixel-Aligned Features for Monocular Human Reconstruction.}
Pixel-aligned reconstruction methods have revolutionized 3D clothed human reconstruction, with PIFu~\cite{pifu2019Saito} introducing implicit functions that map 2D pixel features to 3D space for enhanced detail representation. 
Subsequent approaches like PIFuHD~\cite{pifu-hd2020Saito} and Geo-PIFu~\cite{geo-pifu2020He} further improved reconstruction quality through multi-resolution designs and geometric priors.
While effective for visible details, these methods~\cite{pifu2019Saito, pifu-hd2020Saito, geo-pifu2020He, Monoport2020Li} struggle with complex clothing and poses due to their reliance on visible data only.
Recent advances integrate parametric models through SMPL~\cite{smpl2015Loper} framework~\cite{pamir2022Zheng, Arch2020Huang, Arch++2022He, SeSDF2023Cao, HighFidelity2023Liao} and improve fidelity using normal maps and SDF prediction~\cite{icon2022Xiu, econ2023Xiu, phorhum2022Alldieck, s3f2023Corona, gta2023Zhang, sifu2024Zhang, dif2023Yang}.
However, these methods prioritize quality over speed, resulting in high computational costs that limit their applicability for video streams where reconstruction speed is crucial.

\vspace{0.5em}
\noindent\textbf{Pose Deformation for Monocular Video Reconstruction.}
Leveraging the temporal coherence of human shape from video requires an articulated human model to simulate the natural movement of the human body.
The SMPL parametric 3D body model~\cite{smpl2015Loper} is an articulated human model that contains mesh deformation to adjust the surface mesh (skin and clothing) according to the skeleton's motions, maintaining realistic human contours.
Recent works leveraged the deformation capabilities from SMPL model~\cite{smpl2015Loper} with implicit neural fields (NeRF)~\cite{nerf2020Mildenhall} to enable high-quality dynamic reconstruction~\cite{NeuMan2022Jiang, SelfRecon2022Jiang, HumanNerf2022Weng, InstantAvatar2023Jiang, NeuralBody2021Peng, Danbo2022Su, Anerf2021Su, TransHuman2023Pan}, with Vid2Avatar~\cite{Vid2Avatar2023Guo} using pose-conditioned implicit signed-distance fields for geometry and texture representation.
Despite producing realistic results across various poses, these approaches require minutes or hours of training with relatively slow rendering speeds.
Recent approaches have leveraged Gaussian splatting techniques~\cite{3dgsAvatar2024qian, ExAvatar2024moon, GaussianAvatar2024Hu, hugs2024kocabas, liu24-GVA} for improved rendering efficiency while still relying on offline optimization. 
While these video-based methods produce higher quality results than pixel-aligned approaches, their extensive per-video optimization requirements limit their applicability to pre-recorded videos rather than real-time applications.

\vspace{0.5em}
\noindent\textbf{Real-time 3D clothed human reconstruction.}
Existing approaches to real-time 3D clothed human reconstruction face significant limitations.
Methods like Monoport~\cite{Monoport2020Li} achieve near real-time performance by bypassing explicit 3D reconstruction, while \cite{rtvrendering2023rocco} offers real-time rendering but requires hours for the actual reconstruction.
Multi-view approaches~\cite{multiview12015collet, multiview22021Zhang} require calibrated camera setups unsuitable for in-the-wild scenarios, and template-based methods~\cite{LiveCap2019Habermann, MonoPerfCap2018Xu} depend on pre-scanned templates that limit generalization.
Depth-based techniques~\cite{depth42011newcombe, depth22015Newcombe, depth32022Dong} rely on specialized sensors, while FOF~\cite{fof2022Feng} achieves 30 FPS but lacks texture inference capability.
These limitations underscore the need for methods that balance speed and quality without requiring additional hardware or subject-specific templates.


\section{Proposed Methods}
\label{sec:proposed_methods}
We present \textbf{TemPoFast3D}, a novel plug-and play pipeline for fast 3D clothed human reconstruction from monocular video that combines pixel-aligned features with coordinate mapping in canonical space (See Figure~\ref{fig:main_pipeline}).
Our method introduces temporal propagation strategies through volumetric boundary filtering and visibility-guided sampling to achieve faster performance while maintaining reconstruction quality. 
The pipeline's ``plug-and-play" design allows the feature extraction network $\mathcal{F}$ and query networks $\mathcal{G}_s$, $\mathcal{G}_c$ to be replaced with any SMPL-guided pixel-aligned backbone, enabling easy integration of future improvements.
We first discuss preliminaries in Section~\ref{ssec:preliminary}, then detail our canonical space inference framework in Section~\ref{ssec:canspace_inference}, and finally present our temporal propagation strategy in Section~\ref{ssec:temp_pro}.
As a side benefit from connecting pixel-aligned reconstruction to canonical space, our method can naturally extend to reconstruct multi-view data without any modification will be explained in Section~\ref{ssec:exten_mv}.

\subsection{Preliminary}
\label{ssec:preliminary}

\noindent\textbf{Pixel-aligned Implicit Function} 
\label{sssec:pixel-aligned}
Pixel-aligned Implicit Function (PIFu) \cite{pifu2019Saito} enable volumetric reconstruction from a single image by learning a mapping between 2D pixel features and 3D occupancy. Given an input image $\mathbf{I}$, the method first processes it through a filter network $\mathcal{F}$ to obtain a feature map $h = \mathcal{F}(\mathbf{I})$ that captures spatial and visual information.
For reconstruction, the method samples 3D query points $\mathbf{x} \in \mathbb{R}^3$ in the camera frustum and projects them onto the image plane to obtain relative pixel coordinates $\mathbf{p}_x = (x,y)$ normalized to $[-1,1]$. A shape query network $\mathcal{G}_s$ then predicts occupancy values by combining the projected coordinates with pixel-aligned features:
\begin{equation}
\mathbf{o} = \mathcal{G}_s(h, \mathbf{x}_p), 
\begin{cases} 
  1 & \text{if } \mathbf{x}_p \text{ lies inside the surface} \\
  0 & \text{otherwise}
\end{cases}
\end{equation}
Similarly, a color query network $\mathcal{G}_c$ predicts RGB values $\mathbf{c} \in \mathbb{R}^3$ at query points using the same pixel-aligned features:
\begin{equation}
\mathbf{c} = \mathcal{G}_c(h, \mathbf{x}_p)
\end{equation}

\noindent\textbf{Canonical-Posed Space Transformation.} To establish a mapping between canonical and posed spaces, we leverage the transformation mechanism derived from SMPL~\cite{smpl2015Loper}. The SMPL model defines a parametric function $\mathcal{S}(\boldsymbol{\beta}, \boldsymbol{\theta})$, where $\boldsymbol{\beta} \in \mathbb{R}^{10}$ represents shape parameters and $\boldsymbol{\theta} \in \mathbb{R}^{3\times24}$ defines pose parameters. The transformation between spaces is computed through Linear Blend Skinning:
\begin{equation}
\mathbf{T}_{s} = \sum_{k=1}^K w_{k,i} \mathbf{G}_k(\boldsymbol{\beta}, \boldsymbol{\theta})
\end{equation}
where $w_{k,i}$ are blend weights and $\mathbf{G}_k(\boldsymbol{\beta}, \boldsymbol{\theta})$ are the joint transformation matrices computed from SMPL parameters. The resulting vertices transformation matrices $\mathbf{T}_{s} \in \mathbb{R}^{N \times 4 \times 4}$ enable bidirectional mapping of SMPL vertices:
\begin{equation}
\label{eq:forward_mapping}
\text{Forward mapping:} \quad \mathbf{s}_p = \mathbf{T}_s\mathbf{s}_c
\end{equation}
\begin{equation}
\label{eq:inverse_mapping}
\text{Inverse mapping:} \quad \mathbf{s}_c = \mathbf{T}_s^{-1}\mathbf{s}_p
\end{equation}
These per-vertex transformation matrices are computed once per frame through the SMPL layer and can be transferred to arbitrary coordinates through association with the average of nearest SMPL vertices.

\subsection{Canonical Space Inference and Coordinate Mapping}
\label{ssec:canspace_inference}

Leveraging pixel-aligned reconstruction methods while facilitating temporal information sharing necessitates performing shape inference in canonical space rather than posed space.
This requires establishing a bidirectional mapping between canonical coordinates and posed space where pixel-aligned features are computed.
Given a set of canonical coordinates $\mathbf{x}_c$, we establish correspondence with SMPL canonical vertices $\mathbf{s}_c$ through K nearest neighbor search\footnote{Details on the number of K neighbor are provided in supplementary.}.
Using this correspondence, we transfer the SMPL vertex transformations $\mathbf{T}_s$ to create coordinate transformations $\mathbf{T}_x \in \mathbb{R}^{N \times 4 \times 4}$, effectively extending the SMPL deformation field to arbitrary points in space.
Following Eq.~\ref{eq:forward_mapping}, the transformation to  is then computed as:

\begin{equation}
\label{eq:to_imagespace}
    \mathbf{x}_p = \mathbf{T}_x\mathbf{x}_c
\end{equation}

\begin{figure}[t]
    \centering
    \includegraphics[width=1\columnwidth]{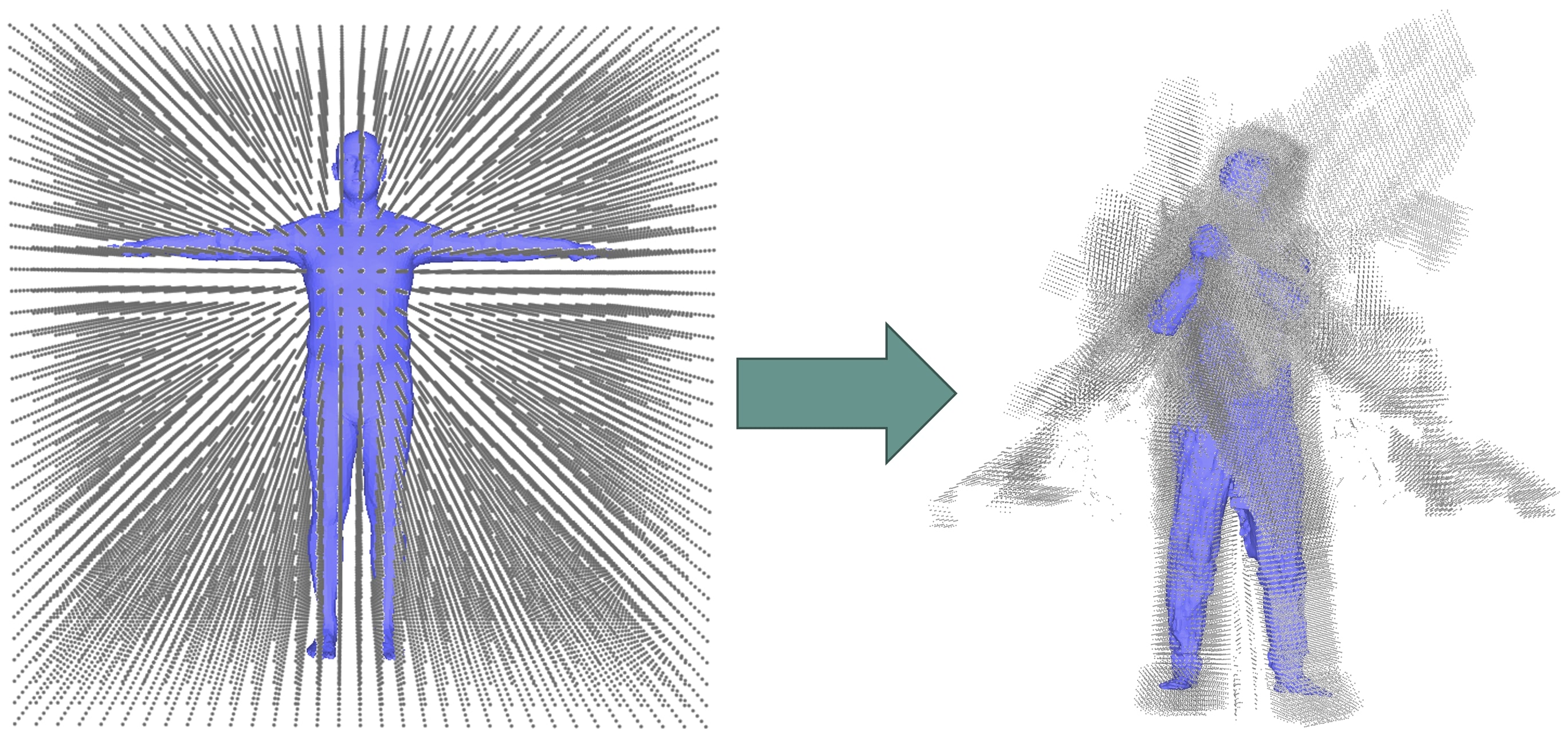}
    \caption{\textbf{Warped sampling coordinate visualization.} Sampling points (gray) transition from uniform distribution in canonical space (left) to non-uniform distribution after deformation (right), demonstrating how our coordinate mapping affects sampling density around the SMPL mesh (blue).}
    \label{fig:warped_sampling}
\end{figure}

Note that the deformation process indiscriminately maps all canonical space coordinates - both inside and outside the body shape - into regions around the deformed body, where the geometric consistency of the reconstruction can be affected by external canonical points being mapped into the interior of the deformed body as illustrated in Figure~\ref{fig:warped_sampling}.

\subsubsection{Volumetric Boundary Filtering}
\label{pg:volumetric}

\begin{figure}[t]
    \centering
    \includegraphics[width=1\columnwidth]{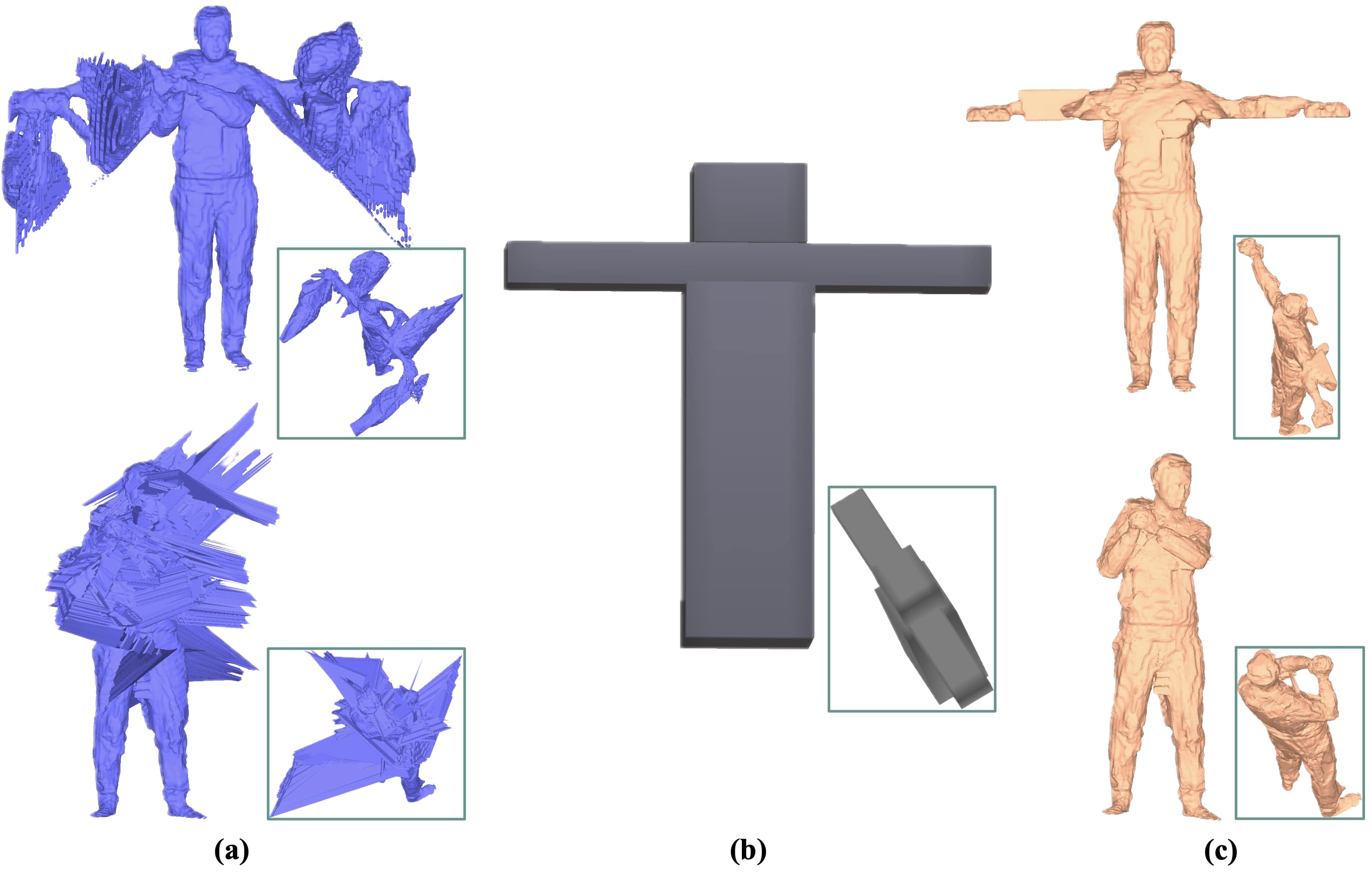}
    \caption{\textbf{Effect of volumetric boundary filtering.} (a) Reconstructed meshes without filtering in canonical (top) and deformed pose (bottom) show artifacts. (b) Volumetric boundary mask. (c) Filtered reconstruction results show cleaner geometry in both poses, eliminating artifacts beyond the valid body region. }
    \label{fig:masking} 
\end{figure}

We observe that valid canonical human geometry predominantly resides within a proximal volume around the canonical SMPL mesh $\mathbf{s}_c $.
A simple discrete volumetric boundary as a mask to this proximal region in canonical space is sufficient to eliminate irrelevant query points.
We formulate this mask as a binary spatial classifier where

\begin{equation}
    \mathbf{m}_i = \begin{cases}
    1 & \text{if } \mathbf{x}_c^i \text{ lies within boundary}\\
    0 & \text{otherwise}
    \end{cases}
\end{equation}
As shown in Figure~\ref{fig:masking}, this filtering mechanism prevents the inclusion of irrelevant canonical coordinates in the shape inference process while additionally reducing the total number of query points required for reconstruction.
\subsubsection{Posed Mesh Generation and Color Inference}
\label{pg:posed}
From the canonical occupancy field, we initially extract an isosurface mesh in canonical pose using the Marching Cubes algorithm.
To obtain the posed configuration, we apply a similar deformation procedure as defined in Eq. \ref{eq:to_imagespace} to transform the mesh into posed space.
Specifically, given canonical vertices $\mathbf{v}_c$, we establish vertex-specific transformation matrices $\mathbf{T}_m$ through nearest neighbor correspondence with canonical SMPL vertices $\mathbf{s}_c$ using K nearest neighbor search, following our earlier coordinate mapping formulation.
The final posed vertices $\mathbf{v}_p$ are then computed by applying these transformations:
\begin{equation}
    \mathbf{v}_p = s(\mathbf{T}_{m}\mathbf{v}_c + \mathbf{t})
\end{equation}
where $s$ and $\mathbf{t}$ are the scale and translation parameters from SMPL estimation.
This transformation enables us to leverage pixel-aligned features for color prediction, as the posed vertices $\mathbf{v}_p$ now align with the input image's space.
Using these pixel-aligned vertices, we employ the color query function $G_c$ from Section~\ref{sssec:pixel-aligned} to predict color values directly from the input image features.
The bidirectional mapping between canonical and posed space enables seamless integration with existing pixel-aligned reconstruction methods for both shape and color inference, while extending the applicability of these methods to temporal sequences.

\subsection{Temporal Propagation and Efficient Inference}
\label{ssec:temp_pro}
While single-frame reconstruction follows the coordinate mapping described in Section~\ref{ssec:canspace_inference}, video sequences offer opportunities for additional computational optimization by leveraging temporal coherency.
We introduce a frame threshold $n$ that determines when to transition from full reconstruction to efficient inference, as illustrated at the top left of Figure~\ref{fig:main_pipeline}. Let $t$ denote the current frame index:
\begin{equation}
\begin{aligned}
&\text{\begin{tabular}{c}
Sampling \\
Strategy
\end{tabular}} = 
\begin{cases} 
\text{Coarse-to-fine sampling} & \text{if } t \leq n, \\
\text{Efficient inference} & \text{if } t >n
\end{cases}
\end{aligned}
\end{equation}
For the initial $n$ frames, we perform complete shape inference to establish a reliable canonical shape representation. After frame $n$, we transition to our efficient inference strategy that enables four key optimizations:

\subsubsection{Bypass Coarse-to-Fine Inference}
\label{pg:bypass}
Temporal propagation strategy establishes a robust geometric prior through propagated canonical shape from previous frames $\mathbf{v}_c^{prev}$, rendering hierarchical coarse-to-fine inference redundant.
This bypass is particularly advantageous in our framework, as each query point incurs additional computational overhead from coordinate mapping operations.
Instead of predicting the entire volume, we focus computation exclusively on regions requiring refinement, reducing the per-frame query complexity.
To identify these regions efficiently, we employ two complementary sampling strategies based on visibility (Section \ref{pg:visibilty}) and surface proximity (Section \ref{pg:surface}).

\begin{table*}[t]
\renewcommand*{\arraystretch}{1.1}
\centering
\resizebox{\textwidth}{!}{%
\begin{tabular}{l|ccc|ccc|cccc}
\toprule
\multirow{2}{*}{Method} & \multicolumn{3}{c|}{CAPE-NFP} & \multicolumn{3}{c|}{CAPE-FP} & \multicolumn{4}{c}{THuman2.0} \\
                        & Chamfer $\downarrow$ & P2S $\downarrow$ & Normal $\downarrow$ & Chamfer $\downarrow$ & P2S $\downarrow$ & Normal $\downarrow$ & Chamfer $\downarrow$ & P2S $\downarrow$ & Normal $\downarrow$ & PSNR $\uparrow$\\
\midrule
                        
PIFu$\ast$~\cite{pifu2019Saito} & 2.5609 & 1.9971  & 0.1023 & 1.8139 & 1.5108  & 0.0798               & 1.5991               & 1.4333          & 0.0843       & 18.09      \\
PIFuHD$\ast$~\cite{pifu-hd2020Saito} & 3.7670 & 3.5910 & 0.1230 & 2.3020  & 2.3350  & 0.0900          & -                    & -               & -  & -        \\ 
ECON$\ast$~\cite{econ2023Xiu}    & 0.9462    & 0.9334          & 0.0382               & 0.9039               & 0.8938          & 0.0373               & 1.2585               & 1.4184          & 0.0612          & -      \\
GTA$\ast$~\cite{gta2023Zhang}   & 0.8508    & 0.7920   & 0.0424   & 0.6525  & 0.6084  & 0.0349               & 0.7329               & 0.7297          & 0.0492            & 18.05   \\
SIFU$\ast$~\cite{sifu2024Zhang}  & \textbf{0.7725}  & 0.7354 & \underline{0.0378}  & 0.6297  & 0.5980 & 0.0327  & 0.5961      & 0.6058 & 0.0407    & 22.10    \\ 
\hline
PIFu$\dagger$~\cite{pifu2019Saito} & 4.2310 & 4.7087  & 0.1029 & 2.5917 & 2.8163 & 0.0827 & 3.1788 & 3.3589  & 0.1082 & -\\  
GTA$\dagger$~\cite{gta2023Zhang} & 0.9160 & 0.8482 & 0.0429 & 0.6531 & 0.6084 & 0.0347 & 0.4625 & 0.4677  & 0.0348 & \textbf{23.27} \\
SIFU$\dagger$~\cite{sifu2024Zhang} & 0.8263 & 0.7889  & 0.0384 & 0.6254 & 0.5901 & 0.0323 & 0.4409 & 0.4580  & 0.0342 & 22.82 \\
\hline
TPF3D-GTA  & 0.9939 & 0.7724  & 0.0507 & 0.7057 & 0.5841  & 0.0383 & 0.5247 & 0.4530  & 0.0383 & \underline{23.25} \\
TPF3D-SIFU & 0.9230 & 0.7147 & 0.0464 & 0.6833 & 0.5663 & 0.0359 & 0.5047 & 0.4432  & 0.0374 & 22.69 \\
TPF3D-GTA-3v   & 0.8293 & \underline{0.6587}  & 0.0391 & \underline{0.5855} & \underline{0.4967} & \underline{0.0285} & \underline{0.4195} &\underline{ 0.3632}  & \textbf{0.0307} & 23.21 \\
TPF3D-SIFU-3v  & \underline{0.8024} & \textbf{0.6351} & \textbf{0.0370} & \textbf{0.5794} & \textbf{0.4883} & \textbf{0.0278} & \textbf{0.4144} & \textbf{0.3590}  & \underline{0.0313} & 22.66 \\
\bottomrule
\end{tabular}}
\caption{\textbf{Quantitative comparison against state-of-the-art methods.} $\ast$: Results of the compared methods obtained from~\cite{sifu2024Zhang}, $\dagger$: We re-evaluated the compared methods for a fair comparison in the same environment (\textit{cf.} Section~\ref{sec:experiments} and Figure~\ref{fig:qualitative}).}
\vspace{-1.5em}
\label{tab:main_results}
\end{table*}

\subsubsection{Visibility-Guided Sampling}
\label{pg:visibilty}
Given a prior canonical shape $\mathbf{v}_c^{prev}$, point queries need only be performed on coordinates that are observable from the current viewpoint, as these regions yield the most reliable predictions from pixel-aligned features.
To identify these regions, we first compute a visibility mask $\mathbf{m}_v$ for the canonical SMPL vertices $\mathbf{s}_c$ through mesh rasterization.
Following our established attribute transfer mechanism (Section~\ref{ssec:canspace_inference}), the visibility status is then propagated to canonical coordinates $\mathbf{x}_c$ through K nearest neighbor search and thus filters out coordinates from unseen region.

\subsubsection{Surface-Adjacent Sampling}
\label{pg:surface}
To achieve better computational efficiency, we constrain point queries to regions near the surface boundary. Specifically, we sample points where the occupancy value $o_c$ falls within a narrow band defined by thresholds $\alpha$ and $\beta$:
\begin{equation}
    \alpha \le o_c \le \beta
\end{equation}
This targeted sampling strategy enables us to maintain detailed surface geometry while significantly reducing query points required for inference.

\subsubsection{Color Propagation and Visibility Handling}
For texture inference, we introduce a visibility-aware color propagation strategy that leverages the canonical space representation to handle occluded regions effectively.
Given a deformed mesh with vertices $\mathbf{v}_d$, we first predict colors for visible vertices $c^{vis}_{v_p}$ using the color query network $\mathcal{G}_c$.
For vertices that are occluded or poorly visible in the current frame, we employ a neighbor-based color propagation scheme that operates in canonical space.
Specifically, we establish correspondence between current canonical vertices $\mathbf{v}_c$ and previous canonical vertices $\mathbf{v}_c^{prev}$ through K nearest neighbor search, thus obtaining color ${c}^{invis}_{v_p}$ from correspondence.
We obtain the final color ${c}_{v_p}$ at time by combining both $c^{vis}_{v_p}$ and ${c}^{invis}_{v_p}$.

\subsection{Extension to Multi-View Inference}
\label{ssec:exten_mv}
The coordinate mapping strategy and canonical inference mechanism (Section~\ref{ssec:canspace_inference}) naturally extend to multi-view reconstruction scenarios, allowing us to aggregate information from synchronized viewpoints without architectural modifications.
By independently processing each view through our pipeline and merging their canonical representations, we leverage the unified canonical space as a consistent global reference for both temporal and spatial fusion.
While not specifically optimized for multi-view scenarios, this capability provides additional validation of our framework's ability to accumulate and refine geometric details through multiple observations.


\section{Experiment}
\label{sec:experiments}

\noindent\textbf{Implementation Details.}
Our pipeline is implemented in PyTorch and executed on a single NVIDIA RTX 4090 GPU.
We utilize PyMAF~\cite{pymaf2021} for SMPL~\cite{smpl2015Loper} parameter regression on in-the-wild data. 
We evaluate our pipeline with GTA~\cite{gta2023Zhang} and SIFU~\cite{sifu2024Zhang} as our pixel-aligned networks (denoted as TPF3D-GTA and TPF3D-SIFU respectively), utilizing their original pre-trained weights to demonstrate our method's plug-and-play capability.
Following the multi-view extension described in Section~\ref{ssec:exten_mv}, we also evaluate three-view configurations (TPF3D-GTA-3v, TPF3D-SIFU-3v) where orthogonal views ({0$^{\circ}$, 120$^{\circ}$, 240$^{\circ}$}) are used to enhance reconstruction quality.
The canonical space reconstruction operates at 256\textsuperscript{3} resolution.
For temporal propagation, we set the frame threshold $n=5$ before transitioning to efficient inference\footnote{Details on hyperparameter \textit{n} are provided in supplementary.}.
Surface-adjacent sampling uses thresholds $\alpha=0.4$ and $\beta=0.7$ to define the sampling region, with random shuffling applied after thresholding for better coverage.

\begin{figure*}[t]
  \centering
   \includegraphics[width=1\linewidth]{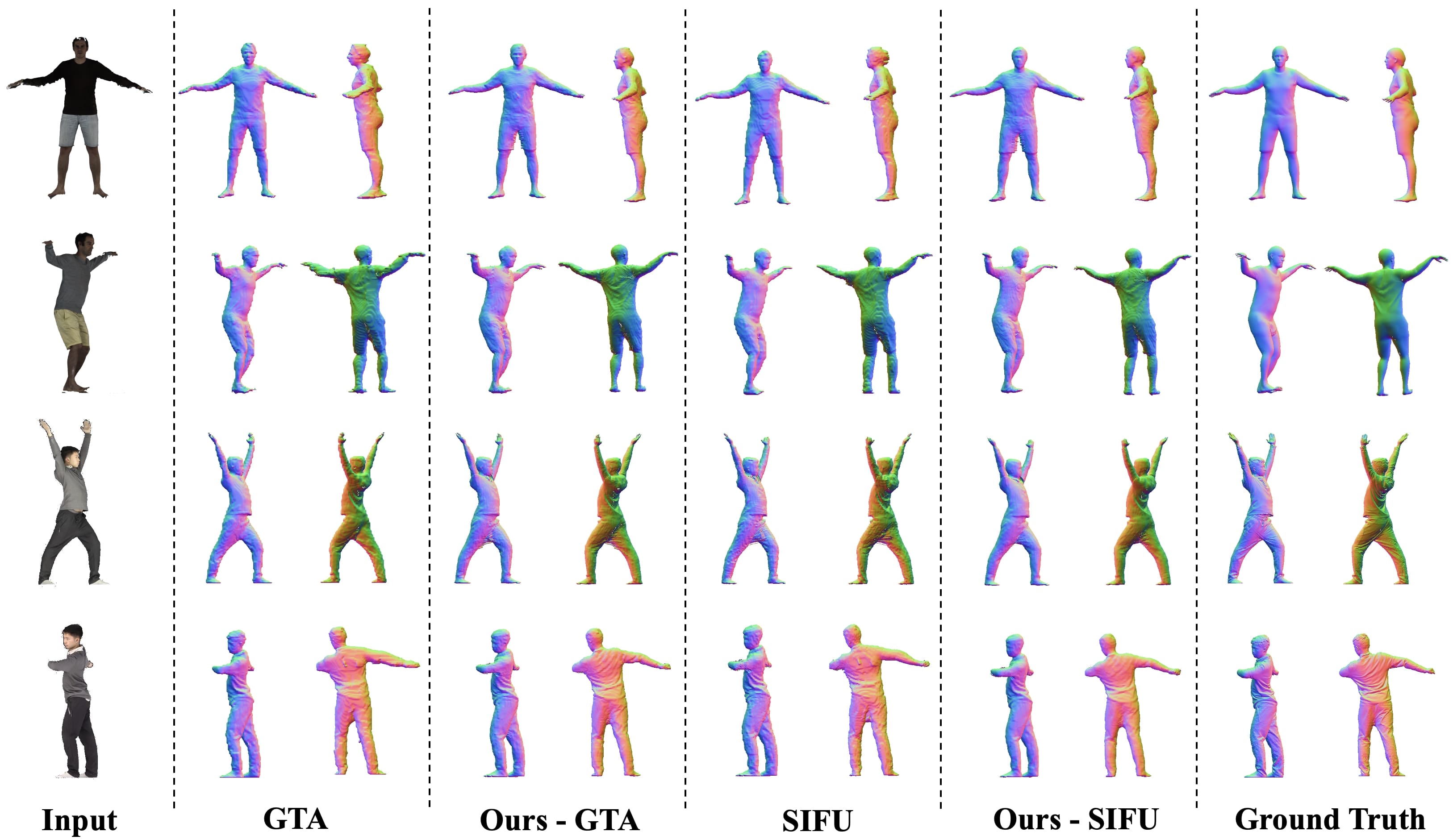}

   \caption{\textbf{Qualitative comparison of geometry reconstruction quality.} The top two rows show results on the CAPE dataset~\cite{CAPEdataset}, while the bottom two rows are from the THuman2.0 dataset~\cite{thuman2dataset}. For best viewing, please zoom in on a digital screen.}
   \label{fig:qualitative}
   \vspace{-1em}
\end{figure*}

\noindent\textbf{Datasets.}
Our pre-trained weight for the pixel-aligned reconstruction networks are trained exclusively on the THuman2.0 dataset~\cite{thuman2dataset}, which consists of 526 human scans along with their corresponding SMPL-X fits.
Of these, 490 are allocated for training, 15 for validation, and 21 for testing.
For zero-shot evaluation, we use the CAPE dataset~\cite{CAPEdataset}. Following previous works, we divide the CAPE dataset into ``CAPE-FP" and ``CAPE-NFP" instead of ``fashion" and ``non-fashion" poses, respectively.
Video performance is evaluated on the NeuMan dataset~\cite{NeuMan2022Jiang}, using the \textit{bike, citron, jogging, and seattle} sequences as per their official testing splits, following previous works~\cite{GaussianAvatar2024Hu, ExAvatar2024moon}.

\noindent\textbf{Evaluation Metrics.}
We employ chamfer distance and P2S (point-to-surface) to evaluate geometric error between ground-truth and predicted mesh.
For reconstruction on single images, shape surface detail and consistency is evaluated using L2 normal error, while texture is evaluated with PSNR.
We utilize a combination of PSNR, SSIM, and LPIPS to evaluate reconstruction accuracy on video data following~\cite{ExAvatar2024moon}.
Quality and speed trade-offs are emphasized by comparing the average FPS for inference/rendering and training time for each video-based method.

\subsection{Evaluation}
\label{ssec:evaluation}

\noindent\textbf{Evaluation on THuman2.0~\cite{thuman2dataset} and CAPE~\cite{CAPEdataset}.}
We first evaluate its performance on single-image datasets to establish baseline capabilities.
Table~\ref{tab:main_results} shows quantitative comparisons with SOTA single-image reconstruction methods on THuman2.0~\cite{thuman2dataset} and CAPE~\cite{CAPEdataset} datasets.
It should be noted that fair comparison is not possible as TPF3D requires multiple view/frame for optimal result while other methods only need single image to achieve max quality.
That being said, Table~\ref{tab:main_results} shows that TPF3D slightly degrades the quality of single-view result but achieves higher reconstruction quality after merging multiple frames ({\it i.e.}, \textbf{TPF3D-SIFU-3v} and \textbf{TPF3D-GTA-3v}).
These scores demonstrate that our approach effectively leverages temporal coherency without sacrificing reconstruction quality.
Figure~\ref{fig:qualitative} presents qualitative comparisons across various poses and clothing styles.
Our method successfully captures fine geometric details such as clothing wrinkles and body contours, particularly in challenging cases like raised arms (row 3) and twisted poses (row 4).
While single-frame methods like GTA~\cite{gta2023Zhang} and SIFU~\cite{sifu2024Zhang} achieve impressive results considering their limited input, our method's ability to combine multiple-view produces a smoother result at the cost of some artifacts due to pose deformation.
For texture reconstruction, our approach achieves comparable PSNR (23.25 dB) to the re-evaluated GTA (23.27 dB).
Additional comparisons are provided in the supplementary material.

\begin{figure*}[t]
  \centering
   \includegraphics[width=0.96\textwidth]{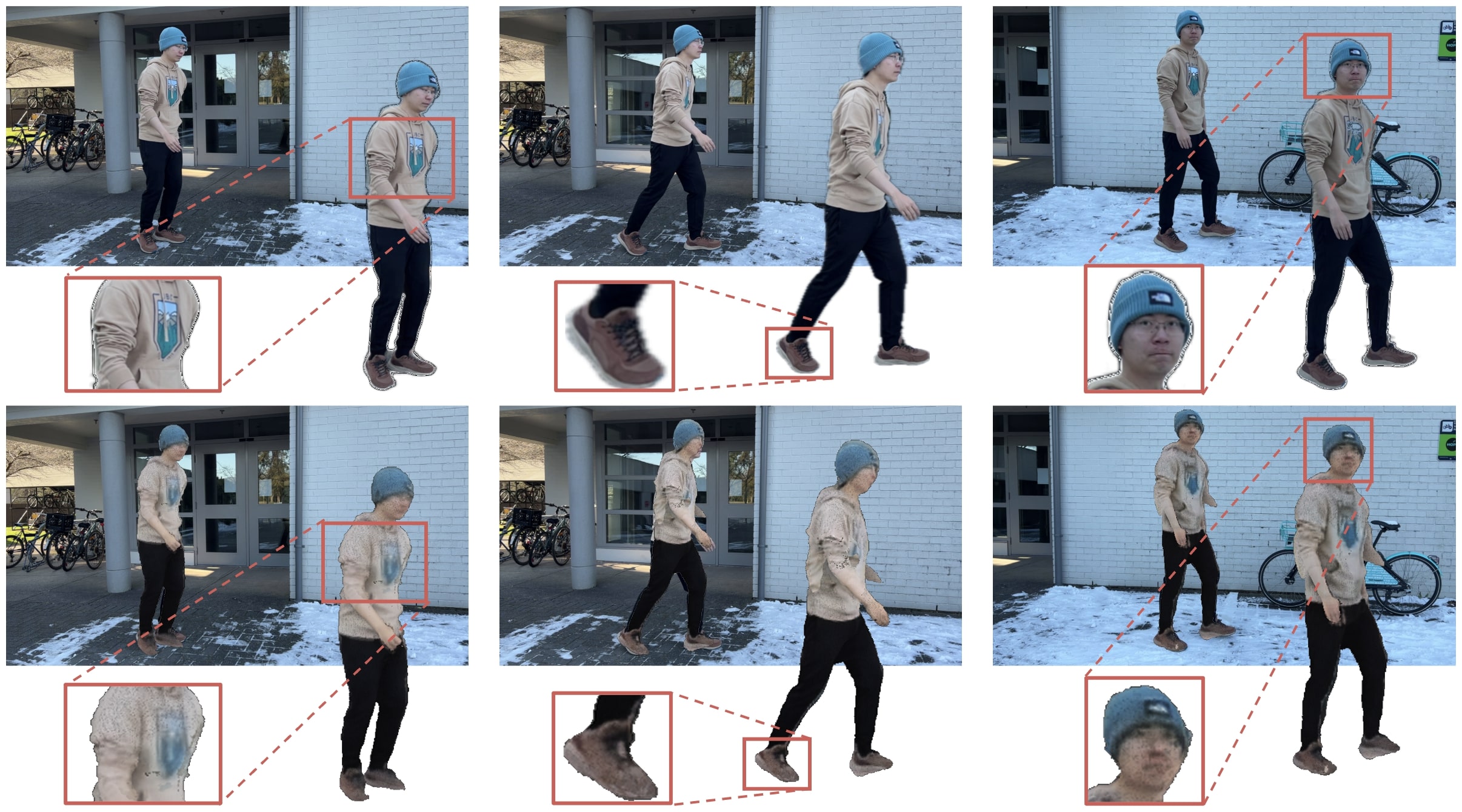}
   \caption{\textbf{Texture quality results on NeuMan~\cite{NeuMan2022Jiang} dataset.} Comparison between ground truth (top) and our real-time reconstruction (bottom) showing consistent quality across early (frame 2) to later frames (frame 22).}
   \label{fig:qualitative_neuman}
\end{figure*}

\noindent\textbf{Evaluation on NeuMan~\cite{NeuMan2022Jiang}.}
We simulate zero-shot inference using in-the-wild videos from NeuMan dataset without their ground truth SMPL parameters.
Similar to evaluation on single-frame reconstruction, fair comparison is not possible as our method does not require per-subject optimization contrary to other methods.
Table~\ref{tab:neuman_eval} shows that our method (TPF3D-GTA) are comparable against early optimization-based approaches in terms of rendering quality (i.e., HumanNerf~\cite{HumanNerf2022Weng}).
ExAvatar~\cite{ExAvatar2024moon} achieves much higher accuracy at the cost of 4 hours of training and slow rendering speed.
GaussianAvatar~\cite{GaussianAvatar2024Hu} and InstantAvatar~\cite{InstantAvatar2023Jiang} achieves faster rendering speed compared to our method, though it should be noted that they require per-subject optimization while our lower FPS includes shape and color reconstruction.
Overall, Table~\ref{tab:neuman_eval} highlights the reconstruction quality and fast speed trade-off for video reconstruction methods, demonstrating that our approach provides a balanced solution with competitive performance and efficient inference.
For qualitative comparison, we compare our reconstruction result with the ground truth in Figure~\ref{fig:qualitative_neuman} for frame 2, 12, and 22 from the \textit{bike} sequences.
TPF3D-GTA method achieves high-quality geometry and texture reconstruction from as early as frame 2, with facial details becoming increasingly clear in later frames.
Minor texture inconsistencies are attributable to our vertex-based color representation.

\begin{table}[t]
\centering
\resizebox{\columnwidth}{!}{%
\begin{tabular}{l | c | c | c | c | c }
\toprule
Method & PSNR $\uparrow$ & SSIM $\uparrow$ & LPIPS $\downarrow$ & Training $\downarrow$ & Avg. FPS $\uparrow$\\
\midrule
    HumanNerf~\cite{HumanNerf2022Weng}	& 27.06	& 0.967	& 0.019 & 26h 33m & 0.540 \\
    InstantAvatar~\cite{InstantAvatar2023Jiang}	& 28.47	& 0.972 & 0.028 & 00h 26m & \textbf{21.000 } \\
    NeuMan~\cite{NeuMan2022Jiang}	& 29.32	& 0.958	& 0.014 & 128h 00m$\ast$ & 0.004 \\
    Vid2Avatar~\cite{Vid2Avatar2023Guo}	& 30.70	& 0.980	& 0.014 &97h 01m & 0.008\\
\hline
    GaussianAvatar~\cite{GaussianAvatar2024Hu}	& 29.94	& 0.980	& 0.012 & 00h 43m & 15.720  \\
    3DGS-Avatar~\cite{3dgsAvatar2024qian}	& 28.99	& 0.974 & 0.016 & - & - \\
    ExAvatar~\cite{ExAvatar2024moon}	& \textbf{34.80} & \textbf{0.984} & \textbf{0.009} & 04h 0m & 4.022 \\
\midrule
\textbf{TPF3D-GTA} (Ours) & 27.60 & 0.965 & 0.022 & \textbf{pretrained} & 8.900 \\
\bottomrule
\end{tabular}}
\caption{\textbf{Quantitative evaluation on NeuMan~\cite{NeuMan2022Jiang}.} PSNR, SSIM, and LPIPS results of other methods are taken from~\cite{ExAvatar2024moon}. 
We run each methods in the same environment\protect\footnotemark for a fair speed comparison, except ``$\ast$" is obtained from~\cite{NeuMan2022Jiang} (\textit{cf.} Section~\ref{ssec:evaluation}).}
\vspace{-1em}
\label{tab:neuman_eval}
\end{table}
\footnotetext{Additional details are provided in supplementary}


\subsection{Ablation Studies}
\label{ssec:ablation}
We evaluate our optimization strategies on the \textit{citron} sequence from the NeuMan dataset, as detailed in Table~\ref{tab:ablation_speed}.
Our baseline implementation achieves 3.27~FPS with PSNR of~32.80 using GTA~\cite{gta2023Zhang}.
Coordinate mapping initially reduces speed, but enables optimizations such as surface-adjacent sampling for a significant speed up, and improved further with limiting the sampling points.
TorchScript optimization provides the final breakthrough, achieving a maximum of 12.3~FPS - a 3.06× speedup over baseline.
Throughout these optimizations, reconstruction quality remains remarkably stable with PSNR above 30.96, SSIM above 0.98, and LPIPS below 0.011, demonstrating that our speed improvements preserve visual fidelity.

\begin{table}[t]
\renewcommand*{\arraystretch}{1.1}
\centering
\resizebox{\columnwidth}{!}{%
\begin{tabular}{l | c | c | c | c }
\toprule
Method & Max. FPS $\uparrow$ & PSNR $\uparrow$ & SSIM $\uparrow$ & LPIPS $\downarrow$ \\
\midrule
Base (GTA~\cite{gta2023Zhang}) & 3.266  & 32.800 & 0.985 & 0.0090 \\
+ Coordinate mapping & 2.142  & 31.024 & 0.983 & 0.0102 \\
+ Linear layer & 2.669  & 30.964 & 0.983 & 0.0102 \\
+ Visibility-guided sampling & 1.908  & 30.965 & 0.983 & 0.0103 \\
+ Surface-adjacent sampling & 4.499  & 30.961 & 0.983 & 0.0102 \\
+ Limit sampling points & 5.841  & 30.998 & 0.983 & 0.0102 \\
+ Torchscript & 12.301  & 31.132 & 0.982 & 0.0106  \\
\bottomrule
\end{tabular}%
}
\caption{\textbf{Ablation study on optimization strategies.} Quantitative comparison of speed (FPS) and quality metrics on NeuMan~\cite{NeuMan2022Jiang}.}
\vspace{-1em}
\label{tab:ablation_speed}
\end{table}

%
%
\section{Conclusion}
We proposed the TemPoFast3D, a novel approach for fast sequential 3D clothed human reconstruction from monocular {RGB} video stream.
We designed the pipeline based on the key idea that human appearance remains largely consistent across video frames, thus complete shape reconstruction every frame was deemed redundant.
TemPoFast3D employed canonical space inference using coordinate mapping to establish a canonical space that can be effectively propagated through time.
This propagated canonical space allowed our pipeline to maintain and refine shape predictions selectively, reducing the computation cost compared to traditional approaches.
Experimental results showed that TemPoFast3D achieves competitive accuracy with state-of-the-art methods while achieving a maximum speed of 12~FPS, pioneering a new direction in efficient 3D human reconstruction through effective temporal information utilization.
Despite so, TemPoFast3D exhibit limitations such as occasional artifacts due to pose deformation as sharp protrusions in select few areas. 
Moreover, our method requires accurate alignment between the predicted mesh and the SMPL model, limiting reconstruction accuracy for loose clothing and accessories.
As a result the reconstruction quality depends on the accuracy of SMPL parameter estimation, which are particularly challenging for extreme poses and occlusions. 
\vspace{-1em}

\section*{Acknowledgment}
This work was partly supported by the Institute of Information \& communications Technology Planning \& Evaluation~(IITP) grant funded by the Korean government~(MSIT) (No.RS-2022-00155915, Artificial Intelligence Convergence Innovation Human Resources Development (Inha University) and  No.RS-2021-II212068, Artificial Intelligence Innovation Hub and IITP-2024-RS-2024-00360227, Leading Generative AI Human Resources Development and No.2022-0-00981, Foreground and Background Matching 3D Object Streaming Technology Development).

{
    \small
    \bibliographystyle{ieeenat_fullname}
    \bibliography{main}
}


\clearpage
\maketitlesupplementary
\setcounter{page}{1}

\section{Implementation Details}

\begin{figure}[t]
    \centering
    \includegraphics[width=\linewidth]{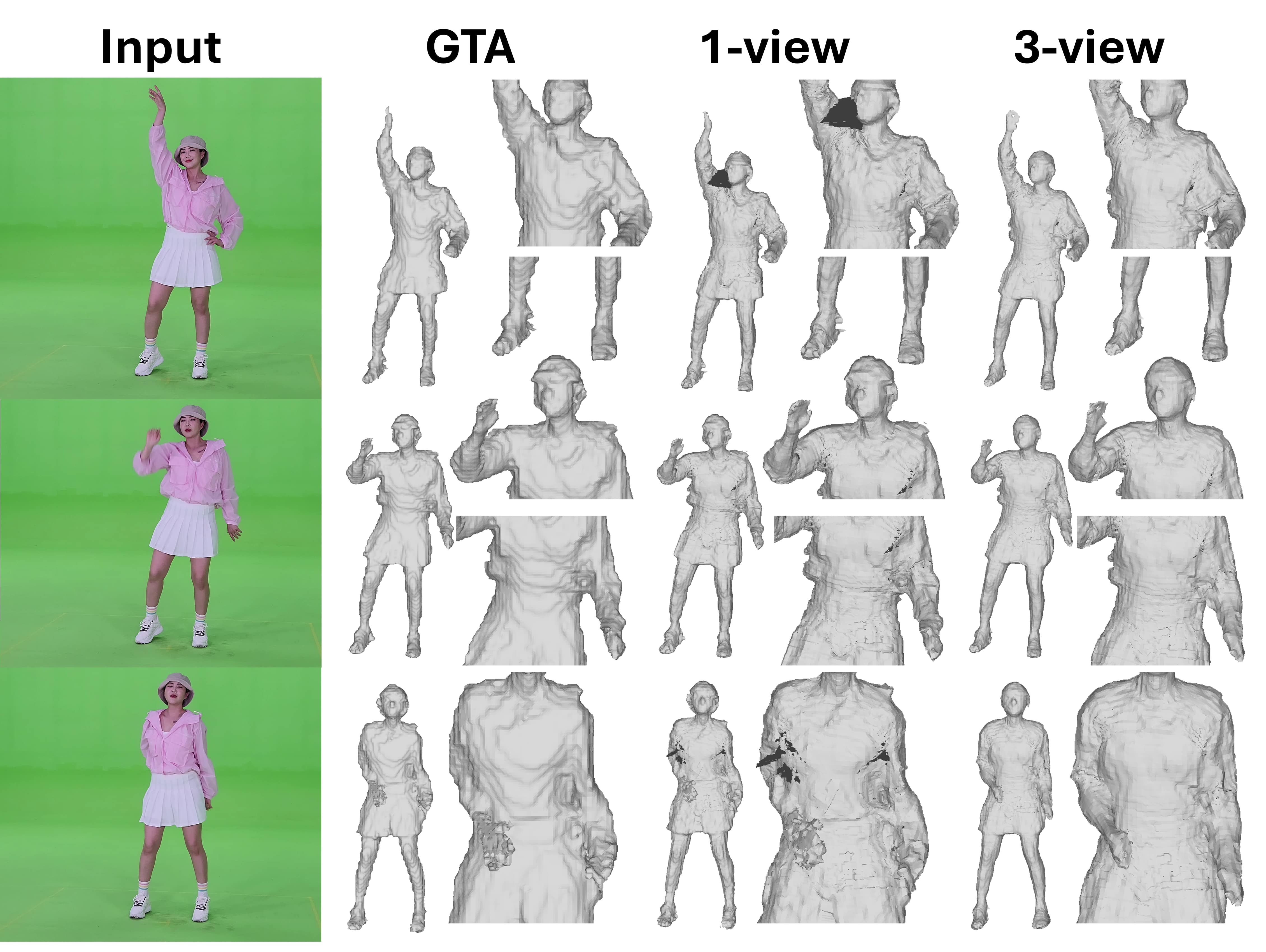}
    \vspace{-0.7cm}
    \caption{Qualitative comparison on clean-background video (for clear comparison) between baseline, single-frame reconstruction, and our three-view approach by merging canonical shapes.}
    \label{fig:comparison_rebuttal} \vspace{-1em}
\end{figure}

\noindent\textbf{Training and Inference Setup.}
All comparative experiments in Table~\ref{tab:neuman_eval} were conducted on a single NVIDIA RTX 4090 GPU.
Training times represent the duration required to train each method only on the ``bike" sequence from NeuMan dataset~\cite{NeuMan2022Jiang}.
Methods without implementation support for NeuMan dataset (e.g., 3DGS-Avatar~\cite{3dgsAvatar2024qian}) were excluded from training time comparison.
We report average FPS across the entire sequence rather than maximum FPS, as our method exhibits speed variation—requiring a slower warm-up period for the first 5 frames (full reconstruction) followed by faster processing for subsequent frames.
Other video-based methods generally maintain consistent processing times throughout the sequence.
In contrast, our ablation studies (Table~\ref{tab:ablation_speed}) report maximum FPS to demonstrate the peak capability of our acceleration strategies after the initial warm-up period.
Only our method's reported FPS in Table~\ref{tab:neuman_eval} includes both shape reconstruction and color inference.

\begin{figure*}[t]
  \centering
   \includegraphics[width=1\linewidth]{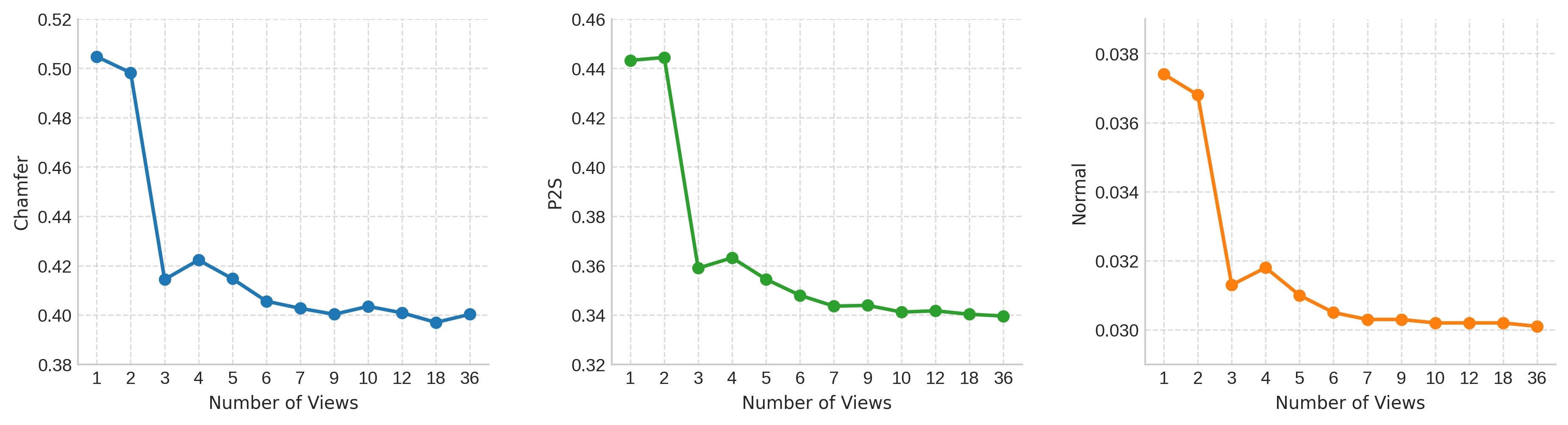}

   \caption{Plotting the results in Table~\ref{tab:num_views} to better visualize the trends in reconstruction quality with respect to number of input views.}
   \label{fig:supp_diff_view}
\end{figure*}

\noindent\textbf{Hyperparameter \textit{n}.}
The frame threshold parameter $n=5$ represents a practical balance between reconstruction quality and computational efficiency.
This value was selected based on our observations during development and is supported by our experimental results in Table \ref{tab:num_views} and Figure \ref{fig:supp_diff_view}, which show diminishing quality returns beyond 6-7 views.
While our multi-view experiments used evenly-spaced orthogonal viewpoints to evaluate the method's theoretical capabilities, the principle of canonical shape convergence applies similarly to sequential frames in video as shown in Figure \ref{fig:comparison_rebuttal}.
The value $n=5$ provides sufficient initial frames to establish a robust canonical human shape while allowing the system to transition to the more efficient inference mode quickly enough to increase the inference speed.
This parameter can be adjusted based on specific application requirements.

\begin{table}[t]
\renewcommand*{\arraystretch}{1.1}
\centering
\resizebox{\linewidth}{!}{%
\begin{tabular}{l|c|ccc}
\toprule
\multirow{2}{*}{Method} & \multirow{2}{*}{K} & \multicolumn{3}{c}{THuman2.0} \\
 & & Chamfer $\downarrow$ & P2S $\downarrow$ & Normal $\downarrow$ \\
\midrule
TPF3D-SIFU & 1  & 0.5253 & 0.4422 & 0.0386   \\
TPF3D-SIFU & 2  & 0.5188 & \textbf{0.4415} & 0.0380   \\
TPF3D-SIFU & 3  & 0.5089 & \underline{0.4420} & \underline{0.0375}   \\
TPF3D-SIFU & 4  & 0.5063 & 0.4421 & \textbf{0.0374}   \\
TPF3D-SIFU & 5  & 0.5048 & 0.4434 & \textbf{0.0374}   \\
TPF3D-SIFU & 6  & \underline{0.5039} & 0.4449 & \underline{0.0375}   \\
TPF3D-SIFU & 7  & \textbf{0.4995} & 0.4461 & 0.0376   \\
\bottomrule
\end{tabular}}
\caption{Comparing the impact of \textbf{K} number of neighbors in coordinate mapping  (Section~\ref{ssec:canspace_inference}) for \textbf{single-frame reconstruction}.}
\label{tab:knn_k_1v}
\end{table}

\begin{table}[t]
\renewcommand*{\arraystretch}{1.1}
\centering
\resizebox{\linewidth}{!}{%
\begin{tabular}{l|c|ccc}
\toprule
\multirow{2}{*}{Method} & \multirow{2}{*}{K} & \multicolumn{3}{c}{THuman2.0} \\
 & & Chamfer $\downarrow$ & P2S $\downarrow$ & Normal $\downarrow$ \\
\midrule
TPF3D-SIFU-3v & 1  & 0.4407 & 0.3596 & 0.0328   \\
TPF3D-SIFU-3v & 2  & 0.4240 & \textbf{0.3576} & 0.0321   \\
TPF3D-SIFU-3v & 3  & 0.4184 & \underline{0.3581} & 0.0315   \\
TPF3D-SIFU-3v & 4  & \underline{0.4162} & \underline{0.3581} & \textbf{0.0313}   \\
TPF3D-SIFU-3v & 5  & \textbf{0.4144} & 0.3590 & \textbf{0.0313}   \\
TPF3D-SIFU-3v & 6  & 0.4179 & 0.3601 & \underline{0.0314}   \\
TPF3D-SIFU-3v & 7  & 0.4182 & 0.3623 & 0.0315   \\
\bottomrule
\end{tabular}}
\caption{Comparing the impact of \textbf{K} number of neighbors in coordinate mapping (Section~\ref{ssec:canspace_inference}) for \textbf{three-frame reconstruction}.}
\label{tab:knn_k_3v}
\end{table}

\begin{figure}[ht!]
  \centering
   \includegraphics[width=0.95\columnwidth]{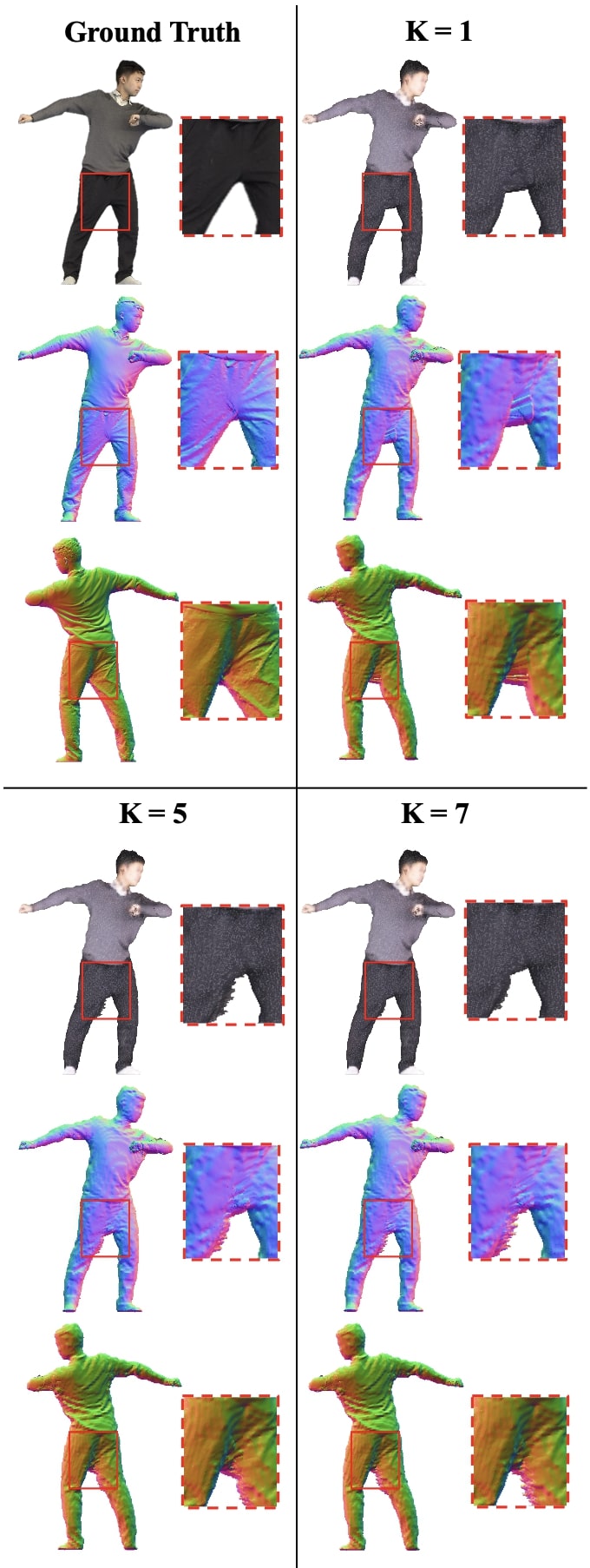}

   \caption{\textbf{Qualitative comparison of geometry reconstruction quality.} under different \textbf{K} values in coordinate mapping.}
   \label{fig:supp_diff_kval_1v}
   \vspace{-2em}
\end{figure}

\noindent\textbf{Impact of K in Coordinate Mapping.}
The number of neighbors (K) in our coordinate mapping affects the trade-off between transformation smoothness and local detail preservation.
We compare the results from single-view reconstruction and three-view reconstruction which we report in Table~\ref{tab:knn_k_1v} and Table~\ref{tab:knn_k_3v}, respectively.
Empirical evaluation shows steady improvement from K=1 to K=5 for chamfer distance and normal consistency while the P2S score decreases.
Larger K values (K $>$ 5) show diminishing returns and eventual degradation in performance.
While these differences are measurable quantitatively, the visual variations in the final reconstruction are subtle, primarily noticeable in the texture creases becoming more defined as K increases, as shown in Figure~\ref{fig:supp_diff_kval_1v}.
We adopt K=5 as our default setting based on these results.

\begin{table}[t]
\renewcommand*{\arraystretch}{1.1}
\centering
\resizebox{\linewidth}{!}{%
\begin{tabular}{l|c|ccc}
\toprule
\multirow{2}{*}{Method} & Num. & \multicolumn{3}{c}{THuman2.0} \\
 & Views & Chamfer $\downarrow$ & P2S $\downarrow$ & Normal $\downarrow$ \\
\midrule
                       
TPF3D-SIFU & 1 & 0.5047 & 0.4432 & 0.0374   \\
TPF3D-SIFU & 2 & 0.4982 & 0.4444 & 0.0368   \\
TPF3D-SIFU & 3 & 0.4144 & 0.3590 & 0.0313   \\
TPF3D-SIFU & 4 & 0.4223 & 0.3632 & 0.0318   \\
TPF3D-SIFU & 5 & 0.4147 & 0.3545 & 0.0310   \\
TPF3D-SIFU & 6 & 0.4056 & 0.3480 & 0.0305   \\
TPF3D-SIFU & 7 & 0.4027 & 0.3436 & 0.0303   \\
TPF3D-SIFU & 9 & 0.4003 & 0.3439 & 0.0303   \\
TPF3D-SIFU & 10 & 0.4035 & 0.3412 & \underline{0.0302}   \\
TPF3D-SIFU & 12 & 0.4009 & 0.3417 & \underline{0.0302}   \\
TPF3D-SIFU & 18 & \textbf{0.3970} & \underline{0.3403} & \underline{0.0302}   \\
TPF3D-SIFU & 36 & \underline{0.4003} & \textbf{0.3396} & \textbf{0.0301}   \\
\bottomrule
\end{tabular}}
\caption{\textbf{Impact of view count on reconstruction quality.} We compare the geometric accuracy improvements by combining results from multiple-views on the THuman2.0 dataset~\cite{thuman2dataset}}
\label{tab:num_views}
\vspace{-1em} 
\end{table}

\begin{figure*}[t]
  \centering
   \includegraphics[width=1\linewidth]{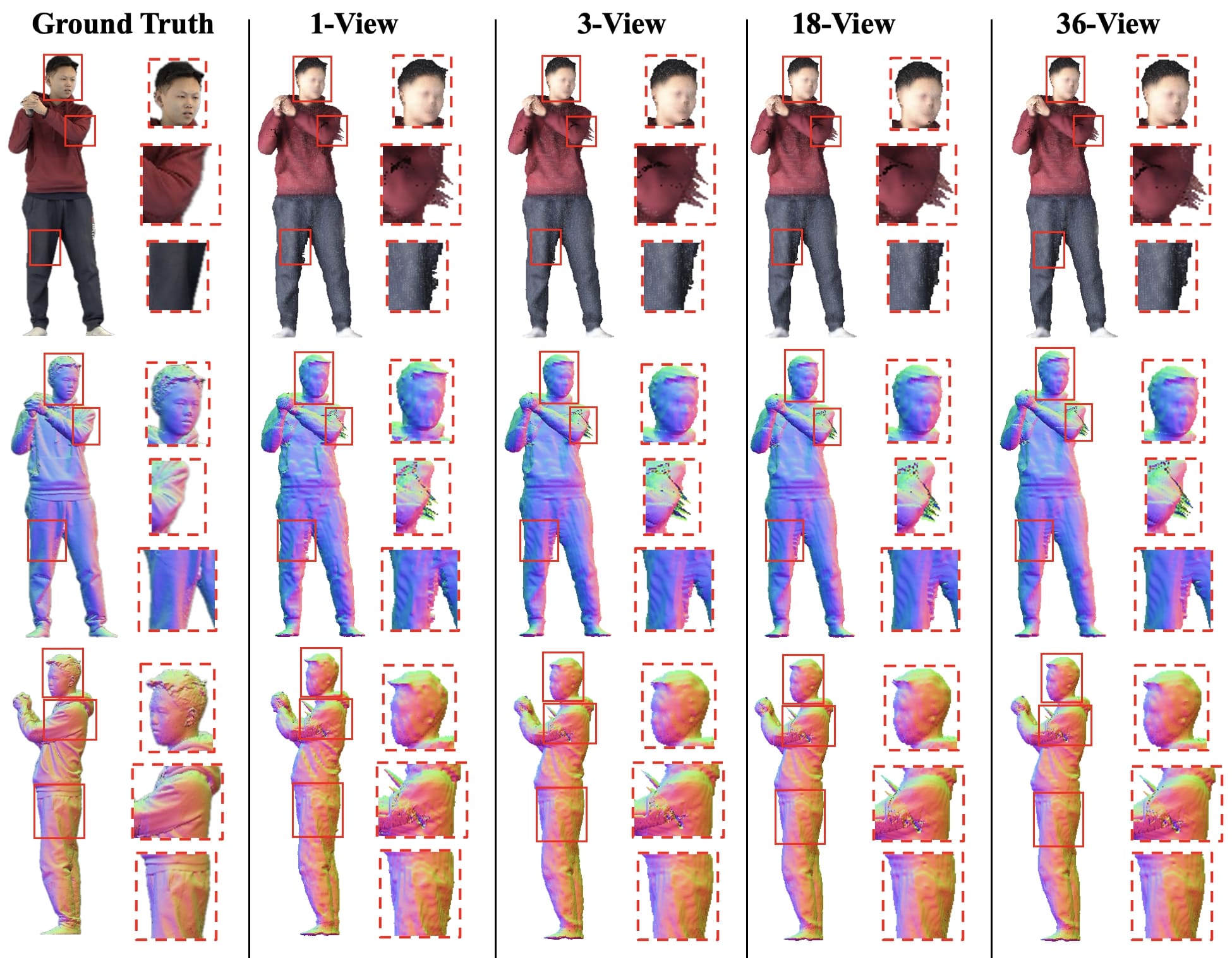}

   \caption{\textbf{Qualitative comparison of geometry reconstruction quality} with varying number of input views}
   \label{fig:supp_diff_view_plot}
\end{figure*}

\vspace{0.5em}\noindent\textbf{Number of views.}
We analyze the relationship between viewpoint multiplicity and reconstruction quality in Table~\ref{tab:num_views}.
Using the THuman2.0~\cite{thuman2dataset} dataset, we evaluate configurations ranging from single to 36-view reconstructions, with viewpoints distributed at maximal angular separations (e.g., {0$^{\circ}$, 180$^{\circ}$} for two views; {0$^{\circ}$, 120$^{\circ}$, 240$^{\circ}$} for three views; {0$^{\circ}$, 90$^{\circ}$, 180$^{\circ}$, 270$^{\circ}$} for four views).
Our analysis reveals consistent improvements in geometric accuracy with additional viewpoints up to 7 views, beyond which returns diminish, ultimately reaching optimal performance at 18 views as illustrated in Figure~\ref{fig:supp_diff_view_plot}.
The qualitative results, visualized in Figure~\ref{fig:supp_diff_view}, validate our multi-view fusion approach while demonstrating the existence of a performance plateau beyond a certain viewpoint threshold.

\begin{figure*}[t]
  \centering
   \includegraphics[width=1\textwidth]{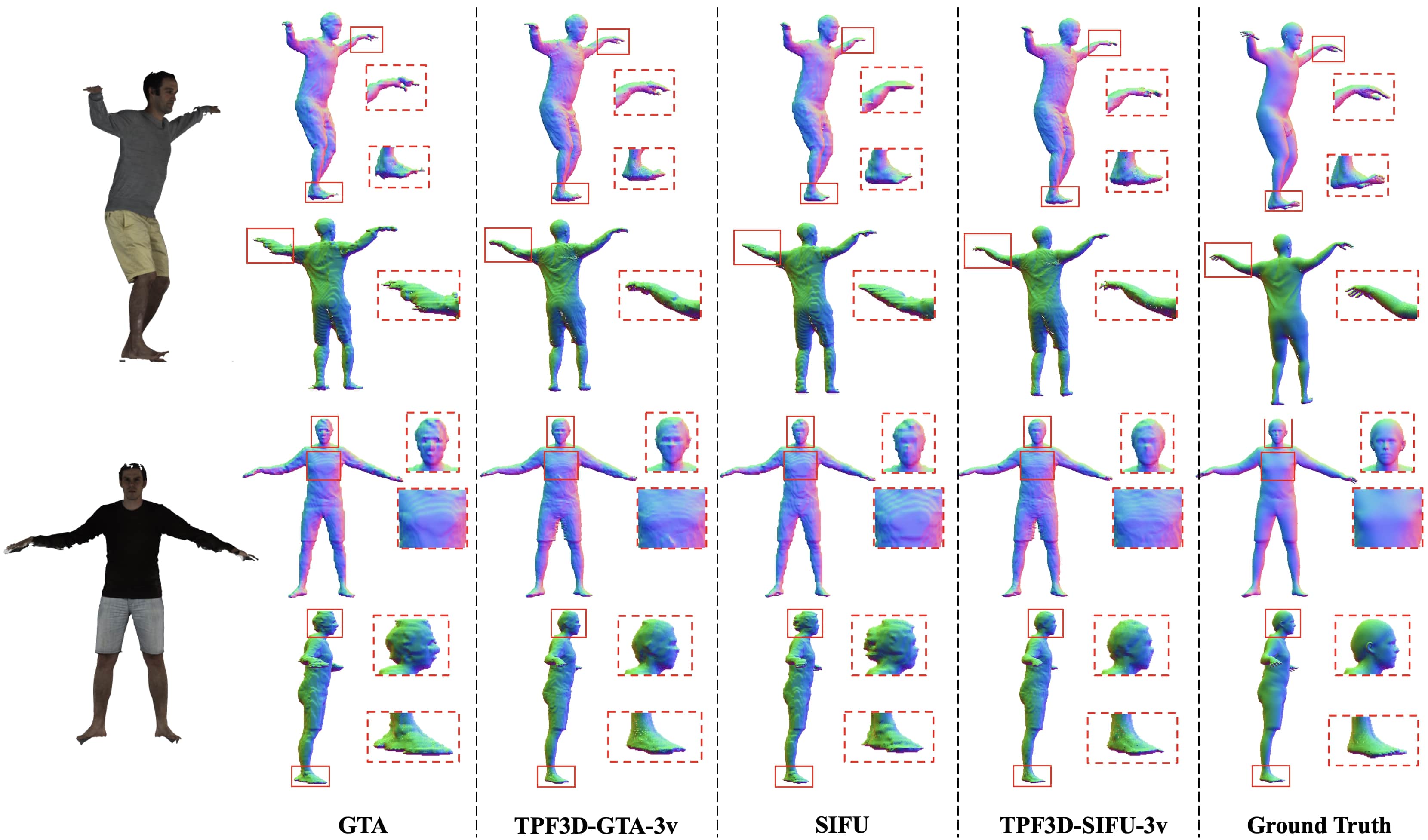}

   \caption{\textbf{Qualitative comparison} of geometry reconstruction quality with state-of-the-art methods. Purple: test view, green: novel view.}
   \label{fig:supp_details_geo}
\end{figure*}

\vspace{-0.5em}
\section{Details on Optimization Strategies}

\noindent\textbf{Baseline.} Our baseline implementation uses GTA~\cite{gta2023Zhang} as the feature extraction backbone, achieving 3.27 FPS while maintaining high reconstruction quality. This represents the unmodified network performing full reconstruction at each frame with uniform sampling across the entire volume.

\vspace{0.5em}\noindent\textbf{Coordinate Mapping.} Introducing coordinate mapping between canonical and posed space initially decreases performance to 2.14 FPS due to the overhead of computing transformation matrices and performing coordinate transformations.
This establishes the foundation for canonical space inference and enables subsequent optimizations for temporal propagation, while temporarily reducing the speed and quality.

\vspace{0.5em}\noindent\textbf{Linear Layer.} We observe that the query networks $\mathcal{G}_s$ and $\mathcal{G}_c$ contain many 1D convolutional layers with $1 \times 1$ filter size, which behave identically to linear layers.
Replacing these with actual linear layers increases speed to 2.67 FPS with minimal quality decrease due to implementation differences between linear and convolutional layers in PyTorch.

\vspace{0.5em}\noindent\textbf{Visibility-Guided and Surface-Adjacent Sampling.} Using visibility-guided sampling alone decreases speed to 1.91 FPS as the number of sampled coordinates remains similar to that of coarse-to-fine inference.
However, combining both visibility-guided and surface-adjacent sampling significantly reduces the coordinate count, increasing performance to 4.50 FPS while maintaining reconstruction accuracy comparable to coordinate mapping.

\vspace{0.5em}\noindent\textbf{Limited Sampling Points.} We further optimize by imposing a strict limit on sampling points (n $<$ 2$^{10}$).
This limit is enforced after the two sampling strategies to ensure points are concentrated in dynamically changing regions. 
As shown in Table~\ref{tab:ablation_speed}, this improves speed to 5.84 FPS without sacrificing quality.

\vspace{0.5em}\noindent\textbf{TorchScript.} The final optimization employs TorchScript compilation to eliminate Python overhead in key computational operations, achieving 3.77$\times$ speedup over baseline (maximum of 12.30 FPS over 3.27 FPS).
This optimization focuses on execution efficiency rather than algorithmic modifications, maintaining reconstruction quality with minimal degradation.

\section{More results}

\begin{figure*}[t]
  \centering
   \includegraphics[width=1\linewidth]{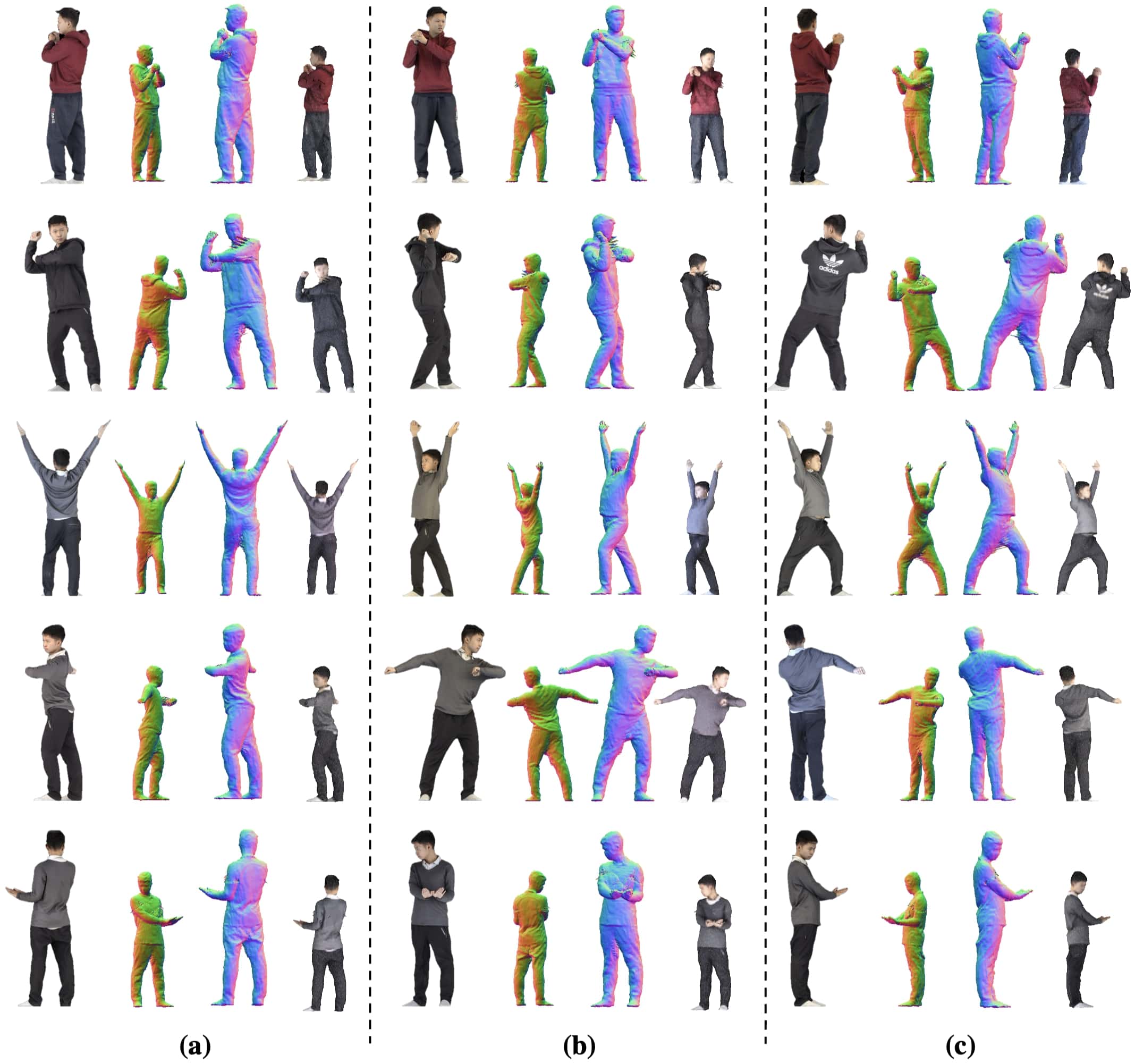}

   \caption{\textbf{Qualitative results} on the THuman2.0~\cite{thuman2dataset} dataset. (a), (b), and (c) represent 0$^{\circ}$, 120$^{\circ}$, and 240$^{\circ}$ test views, respectively. The leftmost column shows the input images, and the rightmost column displays the rendered results on the test view. The purple mesh represents the test view results, while the green mesh corresponds to the novel view results.}
   \label{fig:supp_extra_thuman}
\end{figure*}

We provide additional evaluation results to demonstrate our method's reconstruction capabilities across different scenarios.
In Figure~\ref{fig:supp_details_geo}, we present detailed comparisons with state-of-the-art methods, highlighting the regions with significant differences. Our method (TPF3D-GTA-3v and TPF3D-SIFU-3v) shows improved geometry reconstruction compared to GTA and SIFU baselines. In particular, our approach better preserves fine details in challenging regions such as hands, feet, and head, as shown in the zoomed-in patches. When compared against the ground truth, our reconstructions demonstrate more accurate body proportions and pose estimation, while maintaining geometric details across both test (purple) and novel (green) viewpoints.
Figure~\ref{fig:supp_extra_thuman} showcases comprehensive results on the THuman2.0~\cite{thuman2dataset} dataset, displaying reconstructions from three different angles (0$^{\circ}$, 120$^{\circ}$, 240$^{\circ}$).

\end{document}